\documentclass{article}

\usepackage{arxiv}
\usepackage{float}
\usepackage[nottoc]{tocbibind}

\usepackage[utf8]{inputenc} 
\usepackage[T1]{fontenc}    
\usepackage{hyperref}       
\usepackage{url}            
\usepackage{booktabs}       
\usepackage{amsfonts}       
\usepackage{nicefrac}       
\usepackage{microtype}      
\usepackage{graphicx}
\usepackage[table,xcdraw]{xcolor}
\usepackage{subfig}

\def\mota{MOTA$\uparrow$}
\def\motp{MOTP$\uparrow$}
\def\mt{MT$\uparrow$}
\def\ml{ML$\uparrow$}
\def\frag{F$\downarrow$}
\def\ids{I$\downarrow$}

\def\prec{P$\uparrow$}
\def\rec{R$\uparrow$}

\def\etal{\textit{et~al.}}

\def\absdiff{abs\_diff$\downarrow$}
\def\absrel{abs\_rel$\downarrow$}
\def\sqrel{sq\_rel$\downarrow$}
\def\rms{rms$\downarrow$}
\def\logrms{log\_rms$\downarrow$}
\def\abslog{abs\_log$\downarrow$}
\def\aone{a1$\uparrow$}
\def\atwo{a2$\uparrow$}
\def\athree{a3$\uparrow$}

\title{Virtual KITTI 2}

\author{
  Yohann Cabon, Naila Murray, Martin Humenberger \\
  NAVER LABS Europe \\
  Meylan, France \\
  http://europe.naverlabs.com \\
  \texttt{firstname.lastname@naverlabs.com} \\
}

\begin{document}
\maketitle

\begin{abstract}
This paper introduces an updated version of the well-known Virtual KITTI dataset which consists of 5 sequence clones from the KITTI tracking benchmark. In addition, the dataset provides different variants of these sequences such as modified weather conditions (e.g.~fog, rain) or modified camera configurations (e.g.~rotated by $15^{\circ}$). For each sequence we provide multiple sets of images containing RGB, depth, class segmentation, instance segmentation, flow, and scene flow data. Camera parameters and poses as well as vehicle locations are available as well.
In order to showcase some of the dataset's capabilities, we ran multiple relevant experiments using state-of-the-art algorithms from the field of autonomous driving.
The dataset is available for download at \emph{https://europe.naverlabs.com/Research/Computer-Vision/Proxy-Virtual-Worlds}.
\end{abstract}


\section{Introduction}
Acquiring a large amount of varied and fully annotated data is critical to properly train and test machine learning models for many tasks.
In the past few years, multiple works have demonstrated that while synthetic datasets cannot completely replace real-world data, they are a cost-effective alternative and supplement, and can exhibit good \textit{transferability}~\cite{Gaidon:Virtual:CVPR2016,DeSouza:Procedural:CVPR2017}. For this reason, synthetic datasets can be used to evaluate preliminary prototypes \cite{Gaidon:Virtual:CVPR2016, yang2019drivingstereo} and, sometimes in combination with real-world datasets, to improve performance \cite{DeSouza:Procedural:CVPR2017,Ros_2016_CVPR, yang2019drivingstereo}.

The Virtual KITTI dataset~\cite{Gaidon:Virtual:CVPR2016} was one of the first to explore this approach to training and evaluating models related to driving applications. By carefully recreating real-world videos from the popular KITTI tracking benchmark~\cite{Geiger2012CVPR} in a game engine, \cite{Gaidon:Virtual:CVPR2016} showed that it was possible to generate synthetic data that are comparable to, and can for some applications substitute, real data.

\begin{figure*}[!htb]
\center
\includegraphics[width=1.0\textwidth]{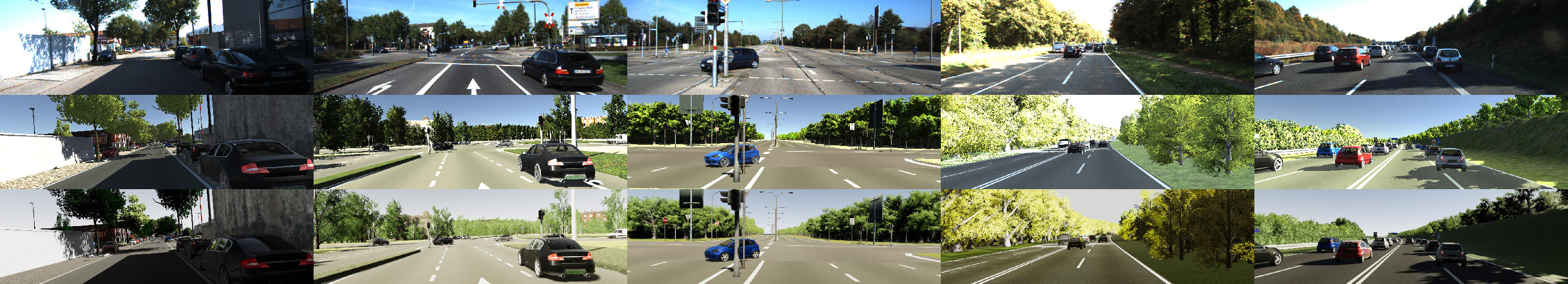}
\vspace*{-4mm}
\caption{\label{fig:pixel_gt} Top row: KITTI images. Middle row: Virtual KITTI 1.3.1 images. Bottom row: Virtual KITTI 2.0 images.} 
\end{figure*}

In this paper, we introduce the new Virtual KITTI 2 dataset. Virtual KITTI 2 is a more photo-realistic and better-featured version of the original virtual KITTI dataset. It exploits recent improvements in lighting and post-processing of the Unity game engine\footnote{https://unity.com/} to bridge the gap between Virtual KITTI and KITTI images (see Figure~\ref{fig:pixel_gt} for examples). 
To showcase the capabilities of Virtual KITTI 2, we re-ran the original experiments of \cite{Gaidon:Virtual:CVPR2016} and added new ones on stereo matching, monocular depth estimation and camera pose estimation as well as semantic segmentation.

\section{Changelog}
The original Virtual KITTI dataset was built using the Unity game engine. It used version 5.3 at time of release then 5.5 for updated versions. It mostly used standard shaders to render objects and no post-processing of any kind was applied to the rendered images.

In Virtual KITTI 2, the scenes were upgraded to Unity 2018.4 LTS and all the materials to \textit{HDRP/Lit} in order to leverage the visual improvements of the High Definition Render Pipeline. The geometry and trajectories were not altered in any way. To increase the application and research field, we added a stereo camera.

To generate the modalities of Virtual KITTI 2, we reused the code we wrote to generate the Virtual Gallery Dataset~\cite{Weinzaepfel:VirtualGallery:CVPR2019} which includes handling of multiple cameras and post-processing of RGB images (supports custom anti-aliasing and the game engine's post processing package). We extended it by adding Virtual KITTI's optical flow implementation.

In all, Virtual KITTI 2 contains the same 5 sequence clones as Virtual KITTI but now provides images from a stereo camera (new). Camera 0 is identical to the Virtual KITTI camera, and camera 1 is 0.532725m to its right. Each camera renders RGB, class segmentation, instance segmentation, depth, forward and backward optical flow (new), and forward and backward scene flow images (new). For each sequence, cameras parameters, vehicle color, pose and bounding boxes are provided.

\section{Related datasets}
In recent years multiple large-scale datasets have been released that focus on furthering appearance and geometry understanding in road scenes. The widely-used Cityscapes~\cite{Cordts2016Cityscapes} dataset contains more than 25k images along with ground-truth semantic segmentation annotations and has been widely used to train and test segmentation models. The SYNTHIA dataset~\cite{Ros_2016_CVPR,HernandezBMVC17,bengarICCVW19} is a synthetic dataset that, similar to Virtual KITTI, was generated using a video game engine. SYNTHIA annotations are focused on semantic segmentation~\cite{Ros_2016_CVPR} and have been successfully used, in combination with real-world data, to improve segmentation model performance. 
DrivingStereo~\cite{yang2019drivingstereo} is a large-scale real-world dataset designed for the stereo matching problem. 
Playing for Benchmarks~\cite{richter2017playing} is another synthetically generated dataset which provides a rich selection of data modalities. 
Virtual Gallery\footnote{https://europe.naverlabs.com/research/3d-vision/virtual-gallery-dataset/}~\cite{Weinzaepfel:VirtualGallery:CVPR2019} is a synthetic dataset generated to target specific challenges of visual localization such as illumination changes or occlusions.
An interesting collection of datasets can be found here\footnote{http://homepages.inf.ed.ac.uk/rbf/CVonline/Imagedbase.htm}.
In a complementary line of work, a method was proposed for validating that a selected dataset is relevant for a given task and contains sufficient variability ~\cite{zendel2017analyzing, zendel2017good}.

Virtual KITTI is unique in that it can be applied to the union of problems addressable by each of the datasets cited above.
More importantly, Virtual KITTI is a reproduction of camera sequences captured in real environments which additionally enables research on domain adaptation, both from sim2real and between different camera angles, filming conditions, and environments~\cite{csurka2017domain}.

\section{Experiments}
\label{sec:experiments}

\subsection{Multi-Object Tracking}
In a first set of experiments, we attempted to reproduce the conclusion in \cite{Gaidon:Virtual:CVPR2016}, which is that in the context of multi-object tracking performance the gap between real and virtual data is small.

In \cite{Gaidon:Virtual:CVPR2016}, \textit{transferability} was tested by comparing multi-object tracking metrics between real and virtual worlds. 
In particular, a pre-trained Fast-RCNN~\cite{girshick2015fast} model was used as a detector and combined with edge boxes proposals~\cite{zitnick2014edge}. 
Two trackers were then evaluated, the dynamic programming min-cost flow (DP-MCF) algorithm  of Pirsiavash~\etal~\cite{pirsiavash2011globally} and the Markov Decision Process (MDP) method of Xiang~\etal~\cite{xiang2015learning}. 
They used Bayesian hyperparameter optimization~\cite{bergstra2013making} to find fixed tracker hyperparameters for each pair of real and clone videos with the sum of the multi-object tracking accuracies~\cite{bernardin2008evaluating} as the objective function.

In this paper, real KITTI videos are compared to Virtual KITTI 1.3.1 and Virtual KITTI 2. Faster-RCNN~\cite{DBLP:journals/corr/RenHG015} (PyTorch implementation~\cite{paszke2017automatic}) replaces Fast-RCNN/Edge Boxes proposals.
ResNet-50 FPN\cite{DBLP:journals/corr/HeZRS15}\cite{DBLP:journals/corr/LinDGHHB16} was used as the backbone. Similar to the original experiment, the network was first trained on imagenet, then fine-tuned on pascal voc 2007~\cite{pascal-voc-2007} and finally on the KITTI object detection benchmark ~\cite{Geiger2012CVPR}. 

The results are reported in table~\ref{tab:resmot}. In this experiment, we show that for every pair there exists a set of DP-MCF parameters for which the MOTA metrics are high and similar between real and virtual data. The conclusions of the original paper therefore still apply.

\begin{table}[H]
\caption{DP-MCF MOT results on original real-world KITTI
train videos and virtual world video ``clones'' (prefixed by a ``v''). real vs 1.31 (left) and real vs 2.0 (right). AVG
(resp. v-AVG) is the average over real (resp.~virtual) sequences. We report the
CLEAR MOT metrics~\cite{bernardin2008evaluating} -- including MOT Accuracy (MOTA), MOT
Precision (MOTP), ID Switches (I), and Fragmentation (F) -- complemented by the
Mostly Tracked (MT) and Mostly Lost (ML) ratios, precision (P) and recall (R).}
\hspace*{-7mm}
\centering
\resizebox{1.07\textwidth}{!}{%
\begin{tabular}{crrrrrrrr}
\rowcolor{gray!50}
DPMCF  &  \mota & \motp  &  \mt    &   \ml &\ids &\frag  & \prec  &  \rec   \\
\toprule                                                                      
real 0001    & 86.3\% & 90.7\% &  92.0\% &  1.3\%&  22 &   43  & 93.6\% & 95.8\% \\
v1.31 0001   & 86.5\% & 81.8\% &  78.9\% &  2.8\%&  10 &   45  & 99.5\% & 89.8\% \\
\midrule                                                                      
real 0002    & 87.3\% & 88.3\% &  90.9\% & 0.0\%&   0 &   6  & 95.6\% & 92.4\% \\
v1.31 0002   & 87.4\% & 81.7\% &  90.0\% & 0.0\%&   1 &   9  & 98.5\% & 89.9\% \\
\midrule                                                                      
real 0006    & 97.3\% & 90.6\% &  100.0\% &  0.0\%&   4 &   5  & 99.7\% & 98.9\% \\
v1.31 0006   & 97.2\% & 88.3\% &  100.0\% &  0.0\%&   2 &   7  & 99.7\% & 98.2\% \\
\midrule                                                                      
real 0018    & 88.6\% & 92.5\% &  47.1\% &  23.5\%&   0 &    2  & 96.5\% & 92.9\% \\
v1.31 0018   & 88.5\% & 69.6\% &  50.0\% &  33.3\%&   0 &    6  & 99.8\% & 89.9\% \\
\midrule                                                                      
real 0020    & 90.9\% & 91.3\% &  87.2\% &  4.7\%&  28 &  45  & 97.3\% & 95.0\% \\
v1.31 0020   & 90.9\% & 79.0\% &  73.4\% &  7.4\%&  5 &  27  & 99.3\% & 92.5\% \\
\midrule                                                                      
real AVG     & 90.1\% & 90.7\% &  83.4\% &  5.9\%&  11 &   20  & 96.5\% & 95.0\% \\
v1.31-AVG   & 90.1\% & 80.1\% &  78.5\% &  8.7\%&  4 &   19  & 99.4\% & 92.1\% \\
\bottomrule
\end{tabular}%
\quad
\begin{tabular}{crrrrrrrr}
\rowcolor{gray!50}
DPMCF   &  \mota & \motp  &  \mt   &   \ml  &\ids &\frag & \prec  &  \rec   \\ 
\toprule                                                                    
real 0001  & 89.9\% & 90.7\% & 93.3\% & 2.7\% &   26 &    44 &  95.0\% & 97.1\% \\
v2.03 0001 & 89.1\% & 84.9\% & 83.1\% & 1.3\% &   12 &    40 &  99.9\% & 91.4\% \\
\midrule                                                                    
real 0002  & 89.8\% & 88.2\% & 90.9\% & 0.0\% &   2 &    7 &  96.5\% & 94.5\% \\
v2.03 0002 & 89.8\% & 82.7\% & 90.0\% & 0.0\% &  1 &    9 &  100.0\% & 90.9\% \\
\midrule                                                                    
real 0006  & 97.3\% & 90.7\% & 100.0\% &  0.0\% &   2 &    3 &  100.0\% & 98.1\% \\
v2.03 0006 & 97.3\% & 86.7\% & 100.0\% &  0.0\% &   3 &    4 &  100.0\% & 98.4\% \\
\midrule                                                                    
real 0018  & 90.9\% & 92.3\% & 88.2\% & 0.0\% &   3 &    4 &  93.4\% & 99.0\% \\
v2.03 0018 & 90.8\% & 75.4\% & 55.6\% & 0.0\% &   4 &    17 &  99.4\% & 92.7\% \\
\midrule                                                                    
real 0020  & 89.4\% & 91.3\% &  86.0\% &  5.8\%&  40 &   69  & 96.7\% & 94.5\% \\
v2.03 0020 & 89.1\% & 83.4\% &  70.0\% &  7.0\%&  19 &   51  & 99.5\% & 91.1\% \\
\midrule                                                                    
real AVG   & 91.5\% & 90.6\% &  91.7\% &  1.7\%&  15 &   25  & 96.3\% & 96.6\% \\
v2.03-AVG & 91.2\% & 82.6\% &  79.7\% &  1.7\%&  8 &   24  & 99.8\% & 92.9\% \\
\bottomrule
\end{tabular}%
}
\vspace*{1mm}
\label{tab:resmot}
\vspace*{-5mm}
\end{table}

For completeness, we also include the results of the variation experiment in table~\ref{tab:resmotvars}. It is noteworthy that the lighting condition changes in Virtual KITTI 2 have a smaller impact on the results. The significantly different implementations of the condition changes make them difficult to compare.

\begin{table}[H]
\caption{Impact of variations on DP-MCF MOT performance in virtual KITTI for the
v1.31 (left) and v2.0 (right) versions.}
\hspace*{-7mm}
\centering
\resizebox{1.07\textwidth}{!}{%
\begin{tabular}{crrrrrrrr}
\rowcolor{gray!50}
v1.31   &  \mota & \motp  &  \mt    &   \ml &\ids &\frag  & \prec  &  \rec    \\
\toprule                                                                         
+15deg   &  -1.3\% & -0.3\% & -2.9\% &  1.5\% & -1 &  -3 &  -0.3\% &  -0.7\% \\
+30deg   &  -3.1\% & -1.5\% & -12.5\% &  4.1\% & -3 &  -9 &  0.3\% &  -2.4\% \\
-15deg   &  -0.5\% & -0.2\% & -2.3\% &  -1.7\% & 1 &  2 &  -0.3\% &  -0.3\% \\
-30deg   &  -3.2\% & -0.6\% & -6.5\% &  1.9\% & 1 &  -1 &  -1.2\% &  -1.8\% \\
morning  &  -6.4\% & -1.3\% &  -7.0\% &  5.0\% &  1 &   0 &  -0.2\% &  -5.5\% \\
sunset   &  -7.5\% & -2.8\% & -9.2\% &  7.2\% &  0 &    2 & -0.9\% &  -6.0\% \\
overcast &  -4.5\% & -2.4\% & -5.8\% &  4.5\% &  1 &   1 &  0.1\% &  -4.0\% \\
fog      & -72.8\% &  3.1\% & -73.8\% & 63.1\% & -3 &  -13 &  0.6\% & -66.7\% \\
rain     &  -29.8\% & -0.8\% & -46.7\% &  17.8\% &  0 &   4 &  0.3\% &  -6.0\% \\
\bottomrule
\end{tabular}%
\quad
\begin{tabular}{crrrrrrrr}
\rowcolor{gray!50}
v2.0   &  \mota & \motp  &  \mt   &   \ml  &\ids &\frag & \prec  &  \rec        \\
\toprule
+15deg   &  -1.1\% & -0.1\% & -6.8\% &  1.7\% & 0 &  -4 &  0.0\% &  -0.7\% \\
+30deg   &  -1.3\% & -1.7\% & -13.3\% &  3.7\% & -5 &  -10 &  -0.1\% &  -0.9\% \\
-15deg   &  -1.1\% & 0.1\% & -0.5\% &  -0.3\% & 4 &  5 &  -0.7\% &  0.5\% \\
-30deg   &  -2.7\% & -0.7\% & -0.6\% &  0.7\% & 3 &  6 &  1.3\% &  -1.4\% \\
morning  &  -0.9\% & -0.3\% &  -4.5\% &  0.7\% &  1 &   4 &  0.1\% &  -0.9\% \\
sunset   &  -2.5\% & -0.5\% & -1.7\% &  1.8\% &  3 &    6 & 0.0\% &  -2.2\% \\
overcast &  -0.3\% & -0.6\% & 2.4\% &  1.2\% &  0 &   3 &  0.0\% &  -0.3\% \\
fog      & -79.6\% &  3.5\% & -78.6\% & 75.5\% & -7 &  -22 &  0.2\% & -73.5\% \\
rain     &  -16.5\% & -0.8\% & -33.0\% &  5.7\% &  -2 &   13 &  0.2\% &  -15.2\% \\
\bottomrule
\end{tabular}%
}
\vspace*{1mm}
\label{tab:resmotvars}
\vspace*{-5mm}
\end{table}

\begin{figure}[H]
\centering
\includegraphics[width=.9\linewidth]{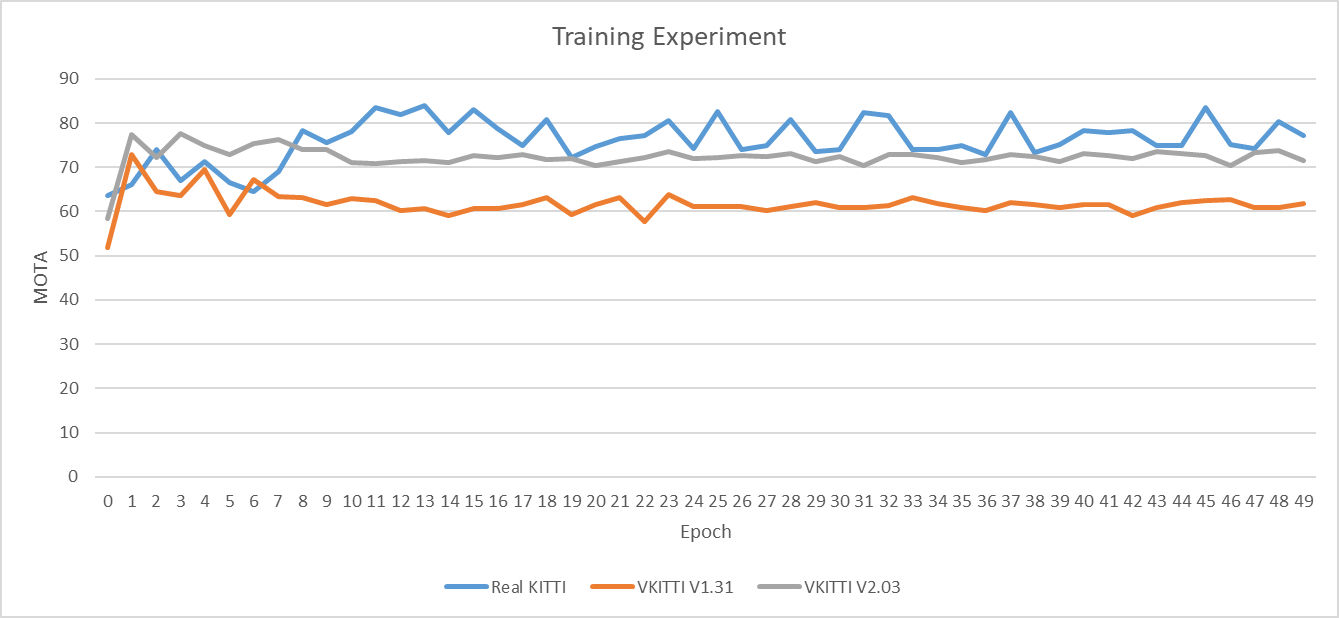}
\caption{Per epoch MOTA results for the training experiment.}
\label{fig:mota_training_chart}
\end{figure}

In the next experiment, we evaluated the training performance of the three datasets. We finetuned the pre-trained imagenet model on the 5 cloned KITTI sequences then tuned the detector on 5 short videos (0,3,10,12,14) and finally evaluated the performance on a test set of 7 long diverse videos (4,5,7,8,9,11,15). 
All three models converged very rapidly (with very few epochs), as shown in Figure~\ref{fig:mota_training_chart}. Detailed metrics are reported in table~\ref{tab:restrain}. In this experiment, the improved RGB seems to have a larger effect closing the gap by 50\%.

\begin{table}[H]
\caption{DP-MCF results on seven long diverse real-world KITTI videos (4,5,7,8,9,11,15) by training the models on the five
cloned KITTI videos (1,2,6,18,20). Reported results correspond to epoch 13.}
\hspace*{-4mm}
\centering
\resizebox{0.60\columnwidth}{!}{%
\begin{tabular}{lrrrrrrrr}
\rowcolor{gray!50}
         &  \mota & \motp  &  \mt  &   \ml &\ids &\frag & \prec  &  \rec  \\
\toprule
real & 83.9\% & 84.3\% & 67.4\% & 15.3\%&   12 &   24& 98.3\% & 88.3\% \\
v1.31 & 60.6\% & 82.1\% & 30.7\% & 30.4\%&   2 &   8& 98.9\% & 64.3\% \\
v2.0 & 71.5\% & 82.9\% & 43.1\% & 18.7\%&   4 &   20& 99.8\% & 75.1\% \\
\bottomrule
\end{tabular}%
}
\vspace*{1mm}
\label{tab:restrain}
\vspace*{-4mm}
\end{table}

\subsection{GANet}

\begin{figure}
\centering
\subfloat[]{\includegraphics[width=.7\linewidth]{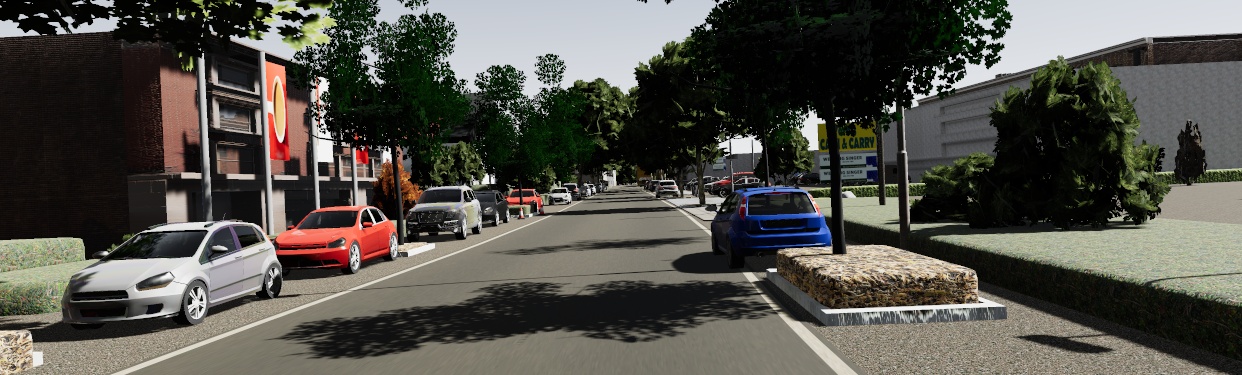}}\\
\subfloat[]{\includegraphics[width=.7\linewidth]{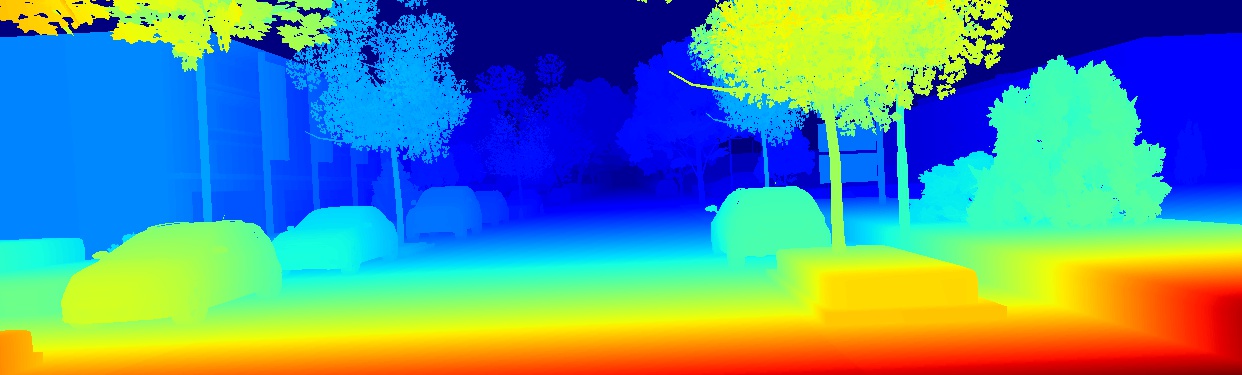}}\\
\subfloat[]{\includegraphics[width=.7\linewidth]{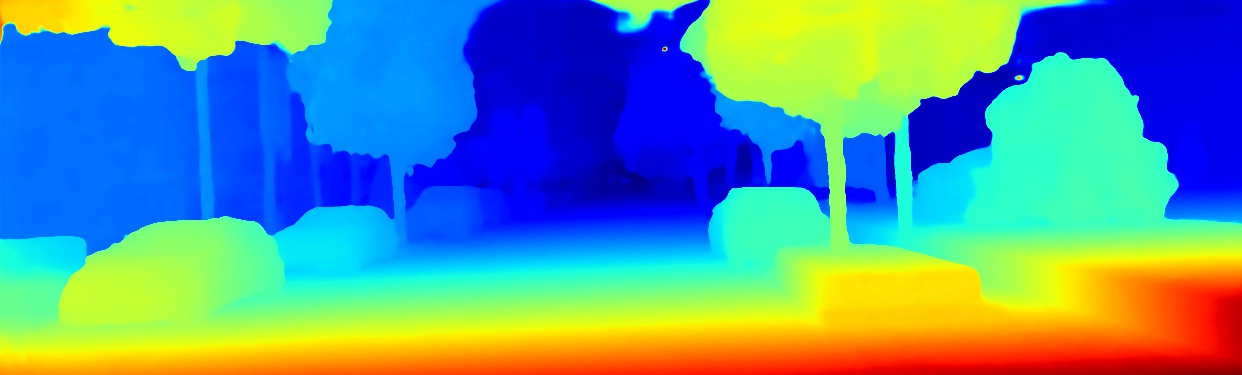}}
\caption{Example image from scene 01: a) RGB image, b) ground-truth disparity map, c) disparity map from GANet (using pre-trained model).}
\label{fig:ganet_dm}
\end{figure}

Since Virtual KITTI 2 also provides stereo image pairs (contrary to the original version), we conducted experiments with a dense deep learning-based stereo matching method, namely GANet~\cite{zhang2019ga}. 
One of the main advantages of Virtual KITTI 1 \& 2 is the ability to test algorithms under different conditions. 
In detail, we can test algorithms with a configuration as close as possible to the real KITTI dataset (clone) as well as with artificial rain, fog or even with slight changes of camera configurations (e.g.~rotated 15 degree left).
In order to show these new capabilities, we ran GANet~\cite{zhang2019ga} on Virtual KITTI 2 as well as on the real KITTI sequences (KITTI tracking benchmark~\cite{Geiger2012CVPR}) using the provided pre-trained models. 
Figure~\ref{fig:ganet_dm} shows one example disparity map (inverse depth) generated using GANet and Virtual KITTI 2. The ground-truth (the reference data provided with our dataset) can be found in the middle and the synthetic camera image can be found at the top.
Tables~\ref{tab:ganet_thresh1_pixel}~and~\ref{tab:ganet_thresh1_percent} show qualitative results of GANet on the real KITTI sequence (real) and Virtual KITTI 2 (all others). There are two main findings: (i) Our clone of the real sequence performs similarly to the real sequence. (ii) Only fog and rain significantly decrease the stereo matching results using this algorithm. Since the original Virtual KITTI does not provide stereo images, we could not run GANet for comparison.

\begin{table}
\centering
\caption{Average error in pixels using the GANet model \emph{kitti2015\_final} and an error threshold of 1 pixel.}
\begin{tabular}{llllll}
\rowcolor{gray!50}
condition & scene01 & scene02 & scene06 & scene18 & scene20 \\
\toprule
real & 1.6421 & 0.7552 & 1.2111 & 2.4188 & 1.5800 \\
clone & 0.9915 & 0.8674 & 0.9201 & 1.7671 & 1.0145 \\
fog & 1.4277 & 1.0595 & 1.0789 & 2.0343 & 1.5217 \\
morning & 1.0374 & 0.8401 & 0.9568 & 1.7454 & 1.3149 \\
overcast & 0.9615 & 0.7892 & 0.9634 & 1.8469 & 1.0638 \\
rain & 1.4500 & 1.2255 & 1.2010 & 1.9652 & 1.3850 \\
sunset & 1.0077 & 0.8336 & 0.9836 & 1.7923 & 1.2003 \\
15-deg-left & 0.9118 & 0.9677 & 0.8344 & 1.4555 & 1.2785 \\
15-deg-right & 1.0643 & 0.9232 & 1.0183 & 2.2778 & 0.7809 \\
30-deg-left & 0.9811 & 0.9439 & 0.8409 & 1.2944 & 1.6015 \\
30-deg-right & 1.2312 & 0.9529 & 0.9061 & 3.0761 & 0.6355 \\
\bottomrule
\end{tabular}
\label{tab:ganet_thresh1_pixel}
\end{table}

\begin{table}
\centering
\caption{Error rate in percent using the GANet model \emph{kitti2015\_final}.}
\begin{tabular}{llllll}
\rowcolor{gray!50}
condition & scene01 & scene02 & scene06 & scene18 & scene20 \\
\toprule
real & 0.1579 & 0.1041 & 0.0853 & 0.2336 & 0.2228 \\
clone & 0.1259 & 0.1420 & 0.1058 & 0.2831 & 0.1427 \\
fog & 0.2010 & 0.2028 & 0.1442 & 0.3257 & 0.2271 \\
morning & 0.1407 & 0.1354 & 0.1048 & 0.2809 & 0.1457 \\
overcast & 0.1208 & 0.1129 & 0.1038 & 0.3027 & 0.1415 \\
rain & 0.1851 & 0.1818 & 0.1806 & 0.3303 & 0.1878 \\
sunset & 0.1324 & 0.1336 & 0.1140 & 0.2871 & 0.1381 \\
15-deg-left & 0.1298 & 0.1448 & 0.0948 & 0.2718 & 0.1675 \\
15-deg-right & 0.1271 & 0.1386 & 0.1096 & 0.3051 & 0.1171 \\
30-deg-left & 0.1436 & 0.1392 & 0.0896 & 0.2717 & 0.1927 \\
30-deg-right & 0.1334 & 0.1426 & 0.1177 & 0.3599 & 0.0964 \\
\bottomrule
\end{tabular}
\label{tab:ganet_thresh1_percent}
\end{table}

\subsection{SfmLearner}

In this experiment, we ran a deep learning-based monocular depth and camera pose estimation algorithm on Virtual KITTI 1 and 2 as well as on the real sequences which were used to generate Virtual KITTI (KITTI tracking benchmark~\cite{Geiger2012CVPR}). We selected a method called SfmLearner~\cite{zhou2017unsupervised} (PyTorch implementation\footnote{https://github.com/ClementPinard/SfmLearner-Pytorch}) and used the pre-trained model provided by the authors for all experiments. Tables~\ref{tab:vkitti_test_disp_scene01}~-~\ref{tab:vkitti2_test_disp_scene02} show the results on depth estimation using SfmLearner on both versions of Virtual KITTI as well as real KITTI (real) on scene 01 and scene 02. 

\begin{table}[H]
\centering
\caption{Results of depth/disparity estimation using the SfmLearner model \emph{kitti\_orig} on scene 01 of Virtual KITTI 1.3.1.}
\begin{tabular}{llllllllll}
\rowcolor{gray!50}
configuration & \absdiff & \absrel & \sqrel & \rms & \logrms & \abslog & \aone & \atwo & \athree \\
\toprule
real & 4.1299 & 0.1871 & 1.6941 & 7.6553 & 0.2969 & 0.1997 & 0.7110 & 0.8765 & 0.9333 \\
clone & 4.3969 & 0.1978 & 1.9045 & 8.5235 & 0.3056 & 0.2086 & 0.6953 & 0.8606 & 0.9453 \\
fog & 7.3598 & 0.3156 & 4.2483 & 13.3553 & 0.5175 & 0.3755 & 0.4597 & 0.7190 & 0.8245 \\
morning & 4.9541 & 0.2196 & 2.2166 & 9.2157 & 0.3326 & 0.2349 & 0.6496 & 0.8322 & 0.9270 \\
overcast & 4.9401 & 0.2216 & 2.1929 & 9.2519 & 0.3331 & 0.2363 & 0.6346 & 0.8355 & 0.9304 \\
rain & 5.3665 & 0.2327 & 2.5394 & 10.2340 & 0.3656 & 0.2554 & 0.6239 & 0.8116 & 0.9097 \\
sunset & 5.0265 & 0.2239 & 2.2690 & 9.4546 & 0.3463 & 0.2430 & 0.6377 & 0.8308 & 0.9248 \\
15-deg-left & 5.5871 & 0.2962 & 3.4072 & 10.0035 & 0.4023 & 0.2855 & 0.5704 & 0.7752 & 0.8869 \\
15-deg-right & 5.2487 & 0.2539 & 2.6596 & 9.6428 & 0.3728 & 0.2622 & 0.6079 & 0.8035 & 0.9111 \\
30-deg-left & 6.3031 & 0.3538 & 4.4495 & 10.9083 & 0.4579 & 0.3333 & 0.5120 & 0.7367 & 0.8495 \\
30-deg-right & 6.2300 & 0.3182 & 3.7206 & 11.1724 & 0.4436 & 0.3209 & 0.5131 & 0.7553 & 0.8674 \\
\bottomrule
\end{tabular}
\label{tab:vkitti_test_disp_scene01}
\end{table}

\begin{table}[H]
\centering
\caption{Results of depth/disparity estimation using the SfmLearner model \emph{kitti\_orig} on scene 01 of Virtual KITTI 2.}
\begin{tabular}{llllllllll}
\rowcolor{gray!50}
configuration & \absdiff & \absrel & \sqrel & \rms & \logrms & \abslog & \aone & \atwo & \athree \\
\toprule
real & 4.1299 & 0.1871 & 1.6941 & 7.6553 & 0.2969 & 0.1997 & 0.7110 & 0.8765 & 0.9333 \\
clone & 4.1806 & 0.2006 & 1.8738 & 8.0914 & 0.2987 & 0.2048 & 0.7053 & 0.8827 & 0.9502 \\
fog & 7.6199 & 0.3590 & 4.6061 & 13.3686 & 0.5464 & 0.4074 & 0.4120 & 0.6748 & 0.8048 \\
morning & 4.7793 & 0.2510 & 2.2305 & 8.3154 & 0.3333 & 0.2479 & 0.5803 & 0.8410 & 0.9386 \\
overcast & 4.8226 & 0.2312 & 2.2000 & 8.9075 & 0.3222 & 0.2352 & 0.6343 & 0.8489 & 0.9336 \\
rain & 6.0536 & 0.2788 & 3.1175 & 10.9652 & 0.4119 & 0.3037 & 0.5267 & 0.7786 & 0.8813 \\
sunset & 4.5673 & 0.2380 & 2.0761 & 8.3144 & 0.3275 & 0.2371 & 0.6190 & 0.8517 & 0.9397 \\
15-deg-left & 4.1439 & 0.1881 & 1.8862 & 8.0820 & 0.2901 & 0.1933 & 0.7233 & 0.8871 & 0.9478 \\
15-deg-right & 4.5730 & 0.2352 & 2.2298 & 8.7058 & 0.3312 & 0.2348 & 0.6383 & 0.8428 & 0.9311 \\
30-deg-left & 4.2628 & 0.2027 & 2.1297 & 8.4016 & 0.3053 & 0.2038 & 0.7037 & 0.8733 & 0.9411 \\
30-deg-right & 4.9444 & 0.2829 & 2.7174 & 9.5778 & 0.3784 & 0.2747 & 0.5799 & 0.7915 & 0.8969 \\
\bottomrule
\end{tabular}
\label{tab:vkitti2_test_disp_scene01}
\end{table}

\begin{table}[H]
\centering
\caption{Results of depth/disparity estimation using the SfmLearner model \emph{kitti\_orig} on scene 02 of Virtual KITTI 1.3.1.}
\begin{tabular}{llllllllll}
\rowcolor{gray!50}
configuration & \absdiff & \absrel & \sqrel & \rms & \logrms & \abslog & \aone & \atwo & \athree \\
\toprule
real & 2.8982 & 0.1080 & 0.9600 & 6.3456 & 0.1931 & 0.1151 & 0.8665 & 0.9522 & 0.9834 \\
clone & 5.1034 & 0.1574 & 2.1665 & 10.6132 & 0.2861 & 0.1800 & 0.7687 & 0.8747 & 0.9415 \\
fog & 8.2950 & 0.2741 & 4.8755 & 15.8891 & 0.5288 & 0.3545 & 0.5552 & 0.7586 & 0.8290 \\
morning & 5.7313 & 0.1835 & 2.6022 & 11.3875 & 0.3225 & 0.2104 & 0.7188 & 0.8452 & 0.9223 \\
overcast & 5.7112 & 0.1867 & 2.6115 & 11.3082 & 0.3241 & 0.2124 & 0.7156 & 0.8454 & 0.9290 \\
rain & 7.2143 & 0.2311 & 3.7951 & 14.0354 & 0.4290 & 0.2856 & 0.6303 & 0.7837 & 0.8638 \\
sunset & 6.0525 & 0.1892 & 2.9137 & 12.2815 & 0.3568 & 0.2260 & 0.7211 & 0.8368 & 0.8997 \\
15-deg-left & 5.9324 & 0.1918 & 2.9228 & 11.8774 & 0.3470 & 0.2179 & 0.7144 & 0.8304 & 0.9073 \\
15-deg-right & 5.7166 & 0.1994 & 2.7859 & 11.0579 & 0.3299 & 0.2159 & 0.7149 & 0.8458 & 0.9230 \\
30-deg-left & 6.4106 & 0.1955 & 3.3892 & 13.1429 & 0.3873 & 0.2349 & 0.7074 & 0.8243 & 0.8826 \\
30-deg-right & 5.7453 & 0.2084 & 2.8659 & 10.9117 & 0.3291 & 0.2195 & 0.6997 & 0.8424 & 0.9266 \\
\bottomrule
\end{tabular}
\label{tab:vkitti_test_disp_scene02}
\end{table}

\begin{table}[H]
\centering
\caption{Results of depth/disparity estimation using the SfmLearner model \emph{kitti\_orig} on scene 02 of Virtual KITTI 2.}
\begin{tabular}{llllllllll}
\rowcolor{gray!50}
configuration & \absdiff & \absrel & \sqrel & \rms & \logrms & \abslog & \aone & \atwo & \athree \\
\toprule
real & 2.8982 & 0.1080 & 0.9600 & 6.3456 & 0.1931 & 0.1151 & 0.8665 & 0.9522 & 0.9834 \\
clone & 6.9450 & 0.2261 & 3.6433 & 13.6930 & 0.4249 & 0.2818 & 0.6278 & 0.7979 & 0.8717 \\
fog & 8.1171 & 0.2749 & 4.7223 & 15.6484 & 0.5196 & 0.3496 & 0.5383 & 0.7695 & 0.8380 \\
morning & 7.0506 & 0.2129 & 3.8568 & 14.3007 & 0.4326 & 0.2720 & 0.6737 & 0.7971 & 0.8675 \\
overcast & 4.9188 & 0.1694 & 2.0306 & 9.8010 & 0.2762 & 0.1839 & 0.7433 & 0.8855 & 0.9570 \\
rain & 6.5200 & 0.2155 & 3.1826 & 12.8214 & 0.3815 & 0.2567 & 0.6521 & 0.8159 & 0.8910 \\
sunset & 5.2766 & 0.1623 & 2.5095 & 11.0582 & 0.3116 & 0.1864 & 0.7549 & 0.8585 & 0.9246 \\
15-deg-left & 6.1858 & 0.2070 & 2.9970 & 12.2804 & 0.3737 & 0.2495 & 0.6721 & 0.8199 & 0.8980 \\
15-deg-right & 7.3299 & 0.2577 & 4.1986 & 14.3245 & 0.4733 & 0.3194 & 0.5724 & 0.7482 & 0.8411 \\
30-deg-left & 5.7392 & 0.1957 & 2.6946 & 11.5449 & 0.3479 & 0.2294 & 0.7038 & 0.8309 & 0.9204 \\
30-deg-right & 6.0436 & 0.2476 & 3.3457 & 12.1962 & 0.4322 & 0.2896 & 0.6127 & 0.7747 & 0.8635 \\
\bottomrule
\end{tabular}
\label{tab:vkitti2_test_disp_scene02}
\end{table}

Tables~\ref{tab:vkitti_test_pose_ate}~and~\ref{tab:vkitti2_test_pose_ate} show the results of camera pose estimation using the metric absolute translation error (ATE) for Virtual KITTI 1.3.1 and 2. The rotation error (RE) for both Virtual KITTI versions is given in tables~\ref{tab:vkitti_test_pose_re}~and~\ref{tab:vkitti2_test_pose_re}. 

\begin{table}[H]
\centering
\caption{Results of camera pose estimation (ATE, absolute translation error, mean (std)) using the SfmLearner model \emph{kitti\_orig} on all scenes of Virtual KITTI 1.3.1.}
\begin{tabular}{llllll}
\rowcolor{gray!50}
condition & scene01 & scene02 & scene06 & scene18 & scene20 \\ 
\toprule
real & 0.0211 (0.0136) & 0.0124 (0.0117) & 0.0045 (0.0064) & 0.0143 (0.0081) & 0.0184 (0.0086) \\
clone & 0.0207 (0.0130) & 0.0125 (0.0118) & 0.0067 (0.0108) & 0.0148 (0.0086) & 0.0204 (0.0118) \\
fog & 0.0263 (0.0170) & 0.0180 (0.0213) & 0.0081 (0.0169) & 0.0213 (0.0165) & 0.0227 (0.0157) \\
morning & 0.0217 (0.0131) & 0.0119 (0.0109) & 0.0062 (0.0093) & 0.0155 (0.0099) & 0.0181 (0.0110) \\
overcast & 0.0215 (0.0133) & 0.0131 (0.0121) & 0.0062 (0.0091) & 0.0160 (0.0108) & 0.0186 (0.0105) \\
rain & 0.0223 (0.0135) & 0.0141 (0.0134) & 0.0059 (0.0093) & 0.0167 (0.0103) & 0.0214 (0.0117) \\
sunset & 0.0212 (0.0127) & 0.0135 (0.0123) & 0.0056 (0.0080) & 0.0155 (0.0096) & 0.0180 (0.0092) \\
15-deg-left & 0.2063 (0.1145) & 0.1377 (0.1438) & 0.0480 (0.0814) & 0.2019 (0.1282) & 0.2315 (0.1167) \\
15-deg-right & 0.1856 (0.0953) & 0.1240 (0.1265) & 0.0532 (0.0885) & 0.1954 (0.1142) & 0.2228 (0.1090) \\
30-deg-left & 0.3832 (0.2078) & 0.2561 (0.2667) & 0.0960 (0.1624) & 0.3810 (0.2315) & 0.4423 (0.2175) \\
30-deg-right & 0.3611 (0.1879) & 0.2464 (0.2541) & 0.0991 (0.1661) & 0.3688 (0.2154) & 0.4309 (0.2115) \\
\bottomrule
\end{tabular}
\label{tab:vkitti_test_pose_ate}
\end{table}

\begin{table}[H]
\centering
\caption{Results of camera pose estimation (ATE, absolute translation error, mean (std)) using the SfmLearner model \emph{kitti\_orig} on all scenes of Virtual KITTI 2.}
\begin{tabular}{llllll}
\rowcolor{gray!50}
condition & scene01 & scene02 & scene06 & scene18 & scene20 \\ 
\toprule
real & 0.0211 (0.0136) & 0.0124 (0.0117) & 0.0045 (0.0064) & 0.0143 (0.0081) & 0.0184 (0.0086) \\
clone & 0.0212 (0.0130) & 0.0123 (0.0112) & 0.0066 (0.0107) & 0.0146 (0.0091) & 0.0197 (0.0114) \\
fog & 0.0253 (0.0163) & 0.0147 (0.0147) & 0.0060 (0.0101) & 0.0204 (0.0151) & 0.0222 (0.0124) \\
morning & 0.0220 (0.0133) & 0.0110 (0.0093) & 0.0056 (0.0082) & 0.0162 (0.0102) & 0.0183 (0.0109) \\
overcast & 0.0216 (0.0129) & 0.0119 (0.0111) & 0.0051 (0.0074) & 0.0168 (0.0106) & 0.0194 (0.0111) \\
rain & 0.0233 (0.0141) & 0.0134 (0.0123) & 0.0058 (0.0087) & 0.0173 (0.0120) & 0.0219 (0.0118) \\
sunset & 0.0207 (0.0128) & 0.0123 (0.0114) & 0.0049 (0.0070) & 0.0163 (0.0111) & 0.0189 (0.0094) \\
15-deg-left & 0.2060 (0.1143) & 0.1368 (0.1424) & 0.0478 (0.0807) & 0.2029 (0.1284) & 0.2309 (0.1169) \\
15-deg-right & 0.1857 (0.0950) & 0.1241 (0.1266) & 0.0531 (0.0885) & 0.1937 (0.1148) & 0.2266 (0.1092) \\
30-deg-left & 0.3826 (0.2075) & 0.2557 (0.2663) & 0.0956 (0.1616) & 0.3829 (0.2313) & 0.4422 (0.2177) \\
30-deg-right & 0.3616 (0.1878) & 0.2454 (0.2531) & 0.0987 (0.1654) & 0.3695 (0.2152) & 0.4358 (0.2133) \\
\bottomrule
\end{tabular}
\label{tab:vkitti2_test_pose_ate}
\end{table}

\begin{table}[H]
\centering
\caption{Results of camera pose estimation (RE, rotation error, mean (std)) using the SfmLearner model \emph{kitti\_orig} on all scenes of Virtual KITTI 1.3.1.}
\begin{tabular}{llllll}
\rowcolor{gray!50}
condition & scene01 & scene02 & scene06 & scene18 & scene20 \\ 
\toprule
real & 0.0028 (0.0017) & 0.0016 (0.0011) & 0.0026 (0.0037) & 0.0018 (0.0009) & 0.0028 (0.0021) \\
clone & 0.0027 (0.0018) & 0.0017 (0.0008) & 0.0026 (0.0026) & 0.0026 (0.0010) & 0.0024 (0.0013) \\
fog & 0.0076 (0.0102) & 0.0141 (0.0152) & 0.0151 (0.0157) & 0.0050 (0.0041) & 0.0083 (0.0066) \\
morning & 0.0028 (0.0018) & 0.0021 (0.0008) & 0.0027 (0.0034) & 0.0017 (0.0007) & 0.0025 (0.0017) \\
overcast & 0.0029 (0.0020) & 0.0019 (0.0008) & 0.0027 (0.0029) & 0.0022 (0.0010) & 0.0021 (0.0015) \\
rain & 0.0035 (0.0029) & 0.0058 (0.0046) & 0.0081 (0.0078) & 0.0022 (0.0011) & 0.0057 (0.0035) \\
sunset & 0.0033 (0.0033) & 0.0020 (0.0008) & 0.0030 (0.0030) & 0.0019 (0.0007) & 0.0023 (0.0015) \\
15-deg-left & 0.0084 (0.0047) & 0.0045 (0.0044) & 0.0036 (0.0044) & 0.0056 (0.0031) & 0.0057 (0.0049) \\
15-deg-right & 0.0086 (0.0059) & 0.0045 (0.0041) & 0.0035 (0.0044) & 0.0072 (0.0023) & 0.0061 (0.0033) \\
30-deg-left & 0.0212 (0.0130) & 0.0105 (0.0117) & 0.0053 (0.0081) & 0.0169 (0.0102) & 0.0123 (0.0116) \\
30-deg-right & 0.0205 (0.0126) & 0.0085 (0.0093) & 0.0070 (0.0083) & 0.0210 (0.0094) & 0.0160 (0.0082) \\
\bottomrule
\end{tabular}
\label{tab:vkitti_test_pose_re}
\end{table}

\begin{table}[H]
\centering
\caption{Results of camera pose estimation (RE, rotation error, mean (std)) using the SfmLearner model \emph{kitti\_orig} on all scenes of Virtual KITTI 2.}
\begin{tabular}{llllll}
\rowcolor{gray!50}
condition & scene01 & scene02 & scene06 & scene18 & scene20 \\ 
\toprule
real & 0.0028 (0.0017) & 0.0016 (0.0011) & 0.0026 (0.0037) & 0.0018 (0.0009) & 0.0028 (0.0021) \\
clone & 0.0027 (0.0018) & 0.0017 (0.0008) & 0.0030 (0.0036) & 0.0020 (0.0008) & 0.0020 (0.0010) \\
fog & 0.0069 (0.0077) & 0.0099 (0.0089) & 0.0110 (0.0100) & 0.0040 (0.0030) & 0.0082 (0.0067) \\
morning & 0.0032 (0.0023) & 0.0019 (0.0009) & 0.0028 (0.0029) & 0.0018 (0.0008) & 0.0023 (0.0020) \\
overcast & 0.0029 (0.0020) & 0.0019 (0.0008) & 0.0027 (0.0023) & 0.0017 (0.0008) & 0.0024 (0.0016) \\
rain & 0.0037 (0.0028) & 0.0064 (0.0058) & 0.0064 (0.0047) & 0.0032 (0.0022) & 0.0060 (0.0032) \\
sunset & 0.0036 (0.0046) & 0.0018 (0.0008) & 0.0025 (0.0027) & 0.0018 (0.0007) & 0.0020 (0.0012) \\
15-deg-left & 0.0086 (0.0051) & 0.0044 (0.0042) & 0.0040 (0.0055) & 0.0055 (0.0027) & 0.0062 (0.0046) \\
15-deg-right & 0.0089 (0.0059) & 0.0046 (0.0045) & 0.0034 (0.0042) & 0.0064 (0.0023) & 0.0054 (0.0033) \\
30-deg-left & 0.0219 (0.0133) & 0.0105 (0.0120) & 0.0056 (0.0088) & 0.0167 (0.0102) & 0.0134 (0.0115) \\
30-deg-right & 0.0214 (0.0139) & 0.0090 (0.0101) & 0.0062 (0.0073) & 0.0204 (0.0102) & 0.0134 (0.0082) \\
\bottomrule
\end{tabular}
\label{tab:vkitti2_test_pose_re}
\end{table}

\subsection{Semantic segmentation}

In this section we use Virtual KITTI 2's ground-truth semantic segmentation annotations to evaluate a state-of-the-art urban scene segmentation method, Adapnet++~\cite{valada19ijcv}, under Virtual KITTI 2 variations. We use the pre-trained models provided by the authors. These models were trained using the CityScapes~\cite{Cordts2016Cityscapes} dataset. 
The models were trained on 11 semantic categories from CityScapes. We evaluated on the following 7 classes (in addition to background) which were the classes common to both Virtual KITTI 2 and CityScapes: sky, building, road, vegetation, pole, car/truck/bus, and traffic sign. We evaluate on both RGB and depth input settings.
While we would have ideally liked to have used models trained on real KITTI data, there currently exists only a small set of 200 KITTI images with segmentation data. 
Further, because these 200 images do not correspond to an image sequence, we do not have a cloned virtual sequence with which to compare.

\begin{figure}
\centering
  \includegraphics[width=.20\linewidth]{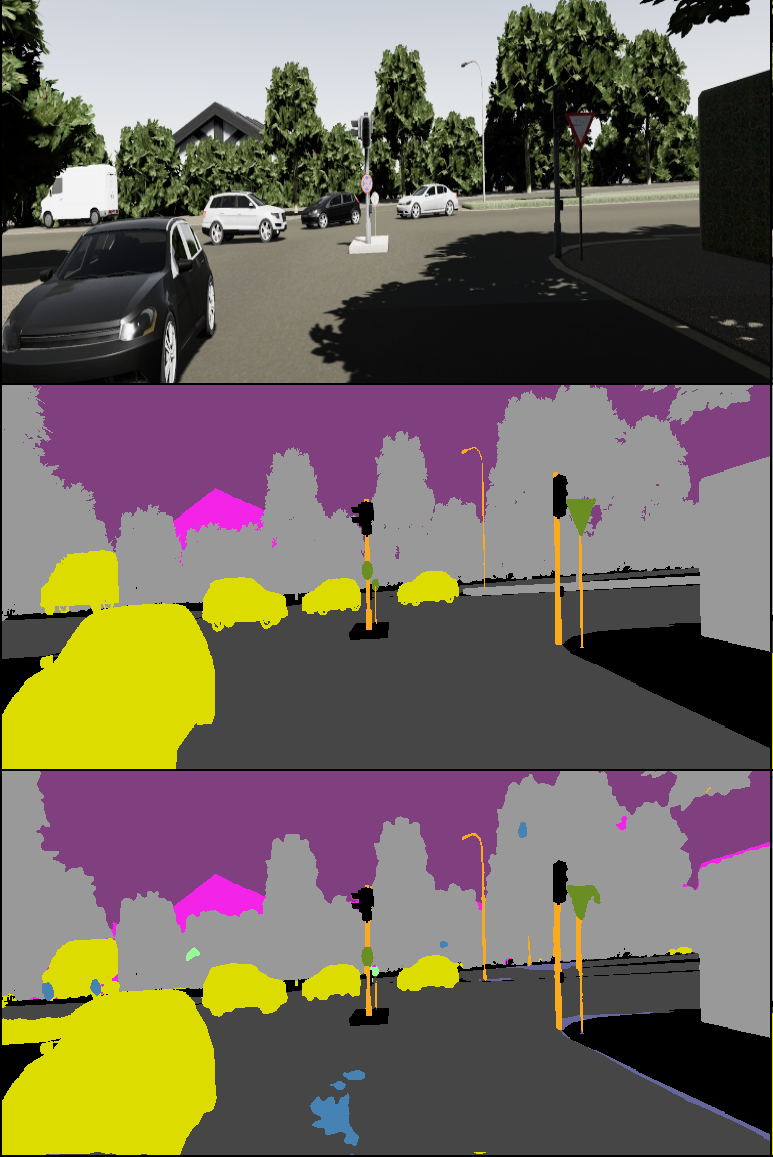}
  \includegraphics[width=.20\linewidth]{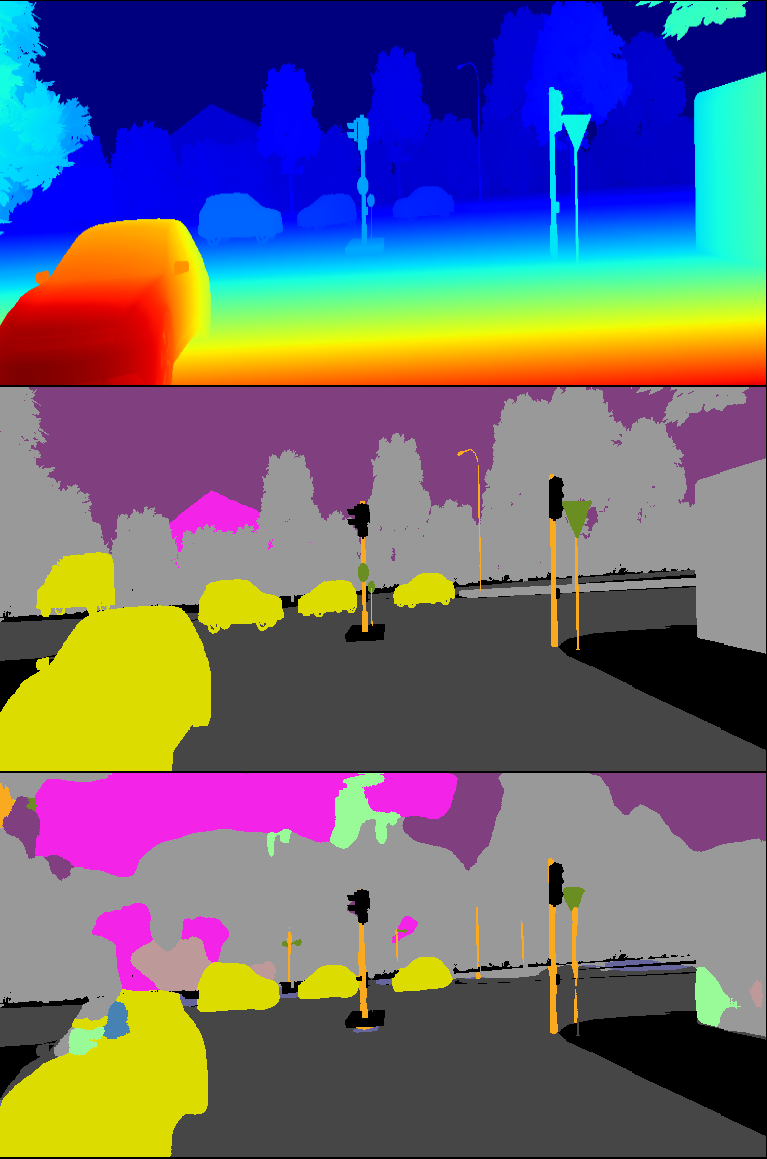}
  \includegraphics[width=.20\linewidth]{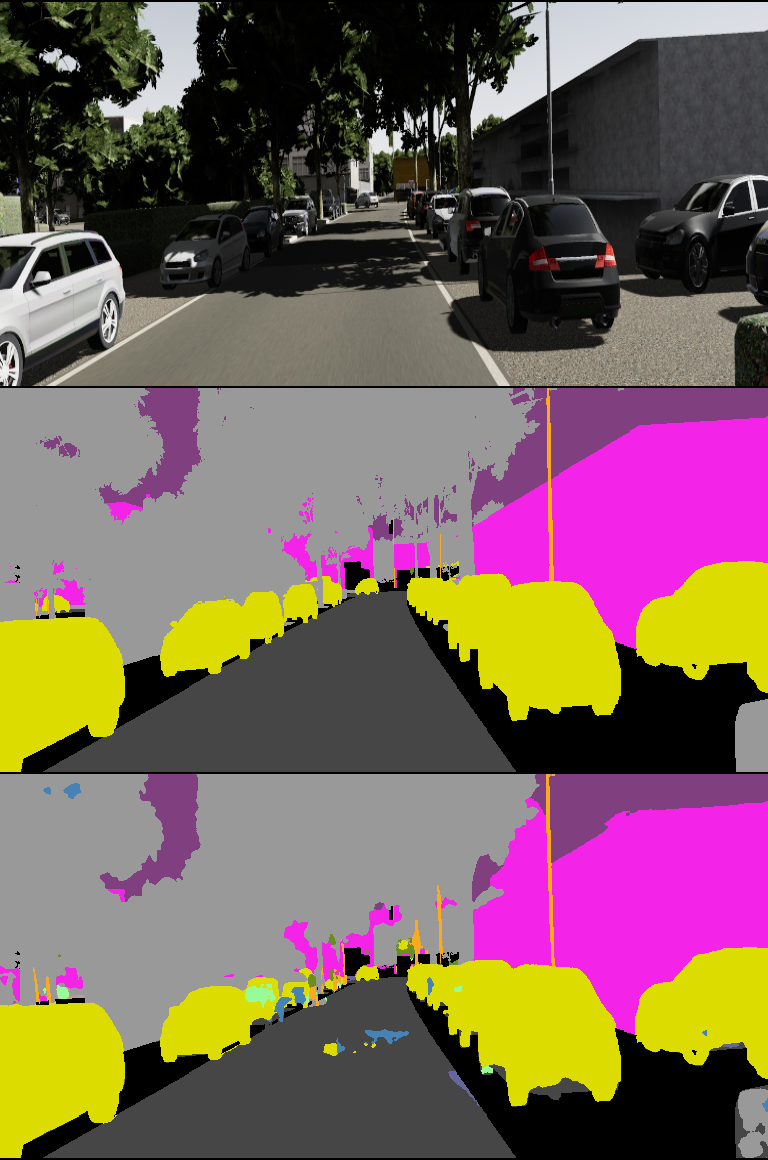}
  \includegraphics[width=.20\linewidth]{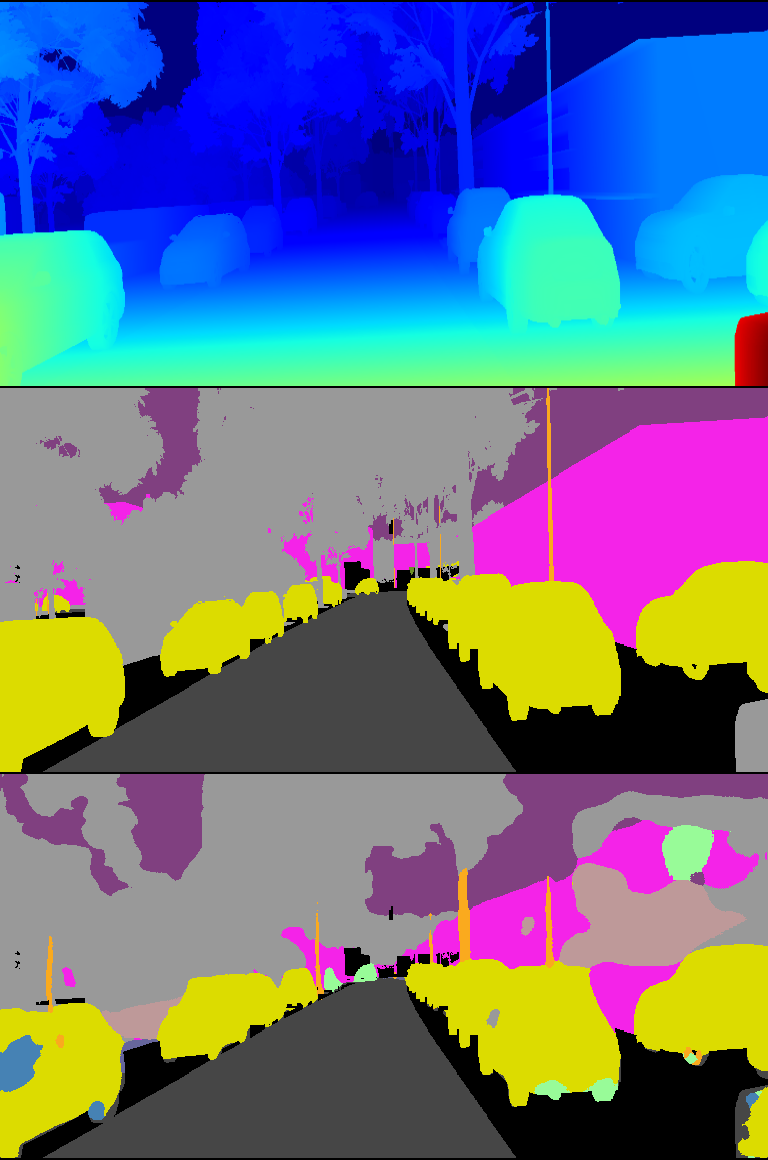}\\
  \includegraphics[width=.20\linewidth]{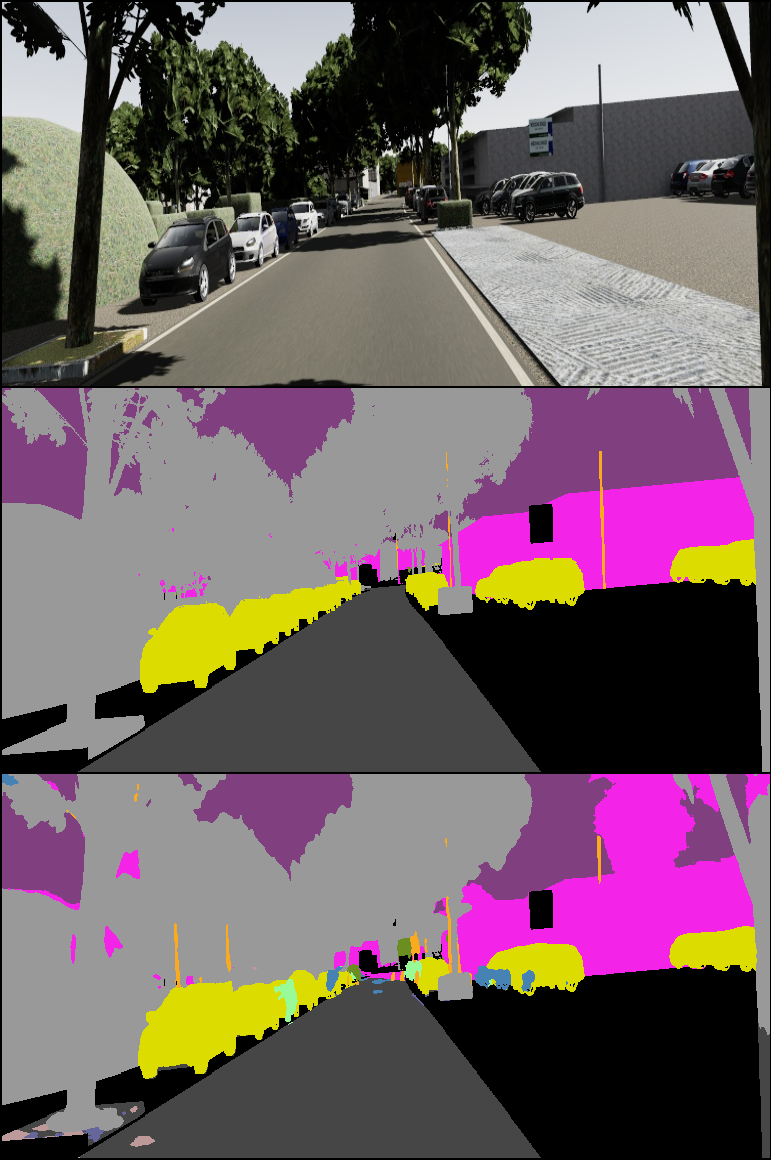}
  \includegraphics[width=.20\linewidth]{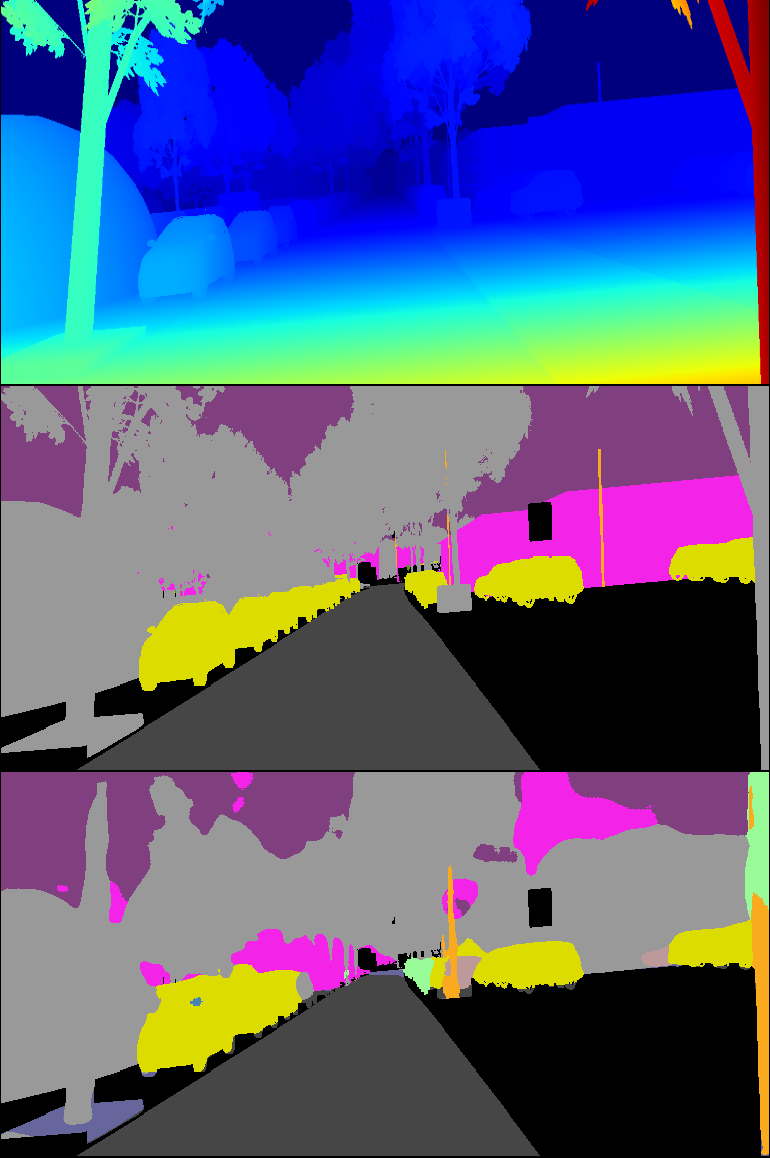}
  \includegraphics[width=.20\linewidth]{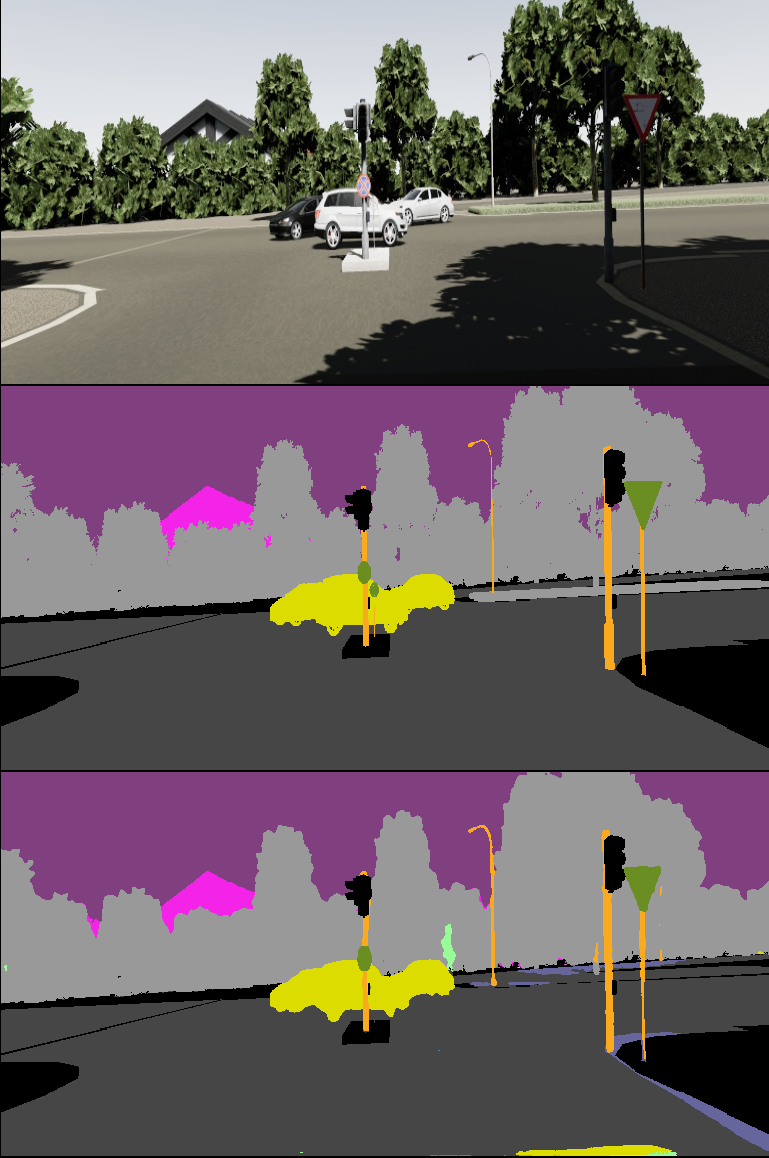}
  \includegraphics[width=.20\linewidth]{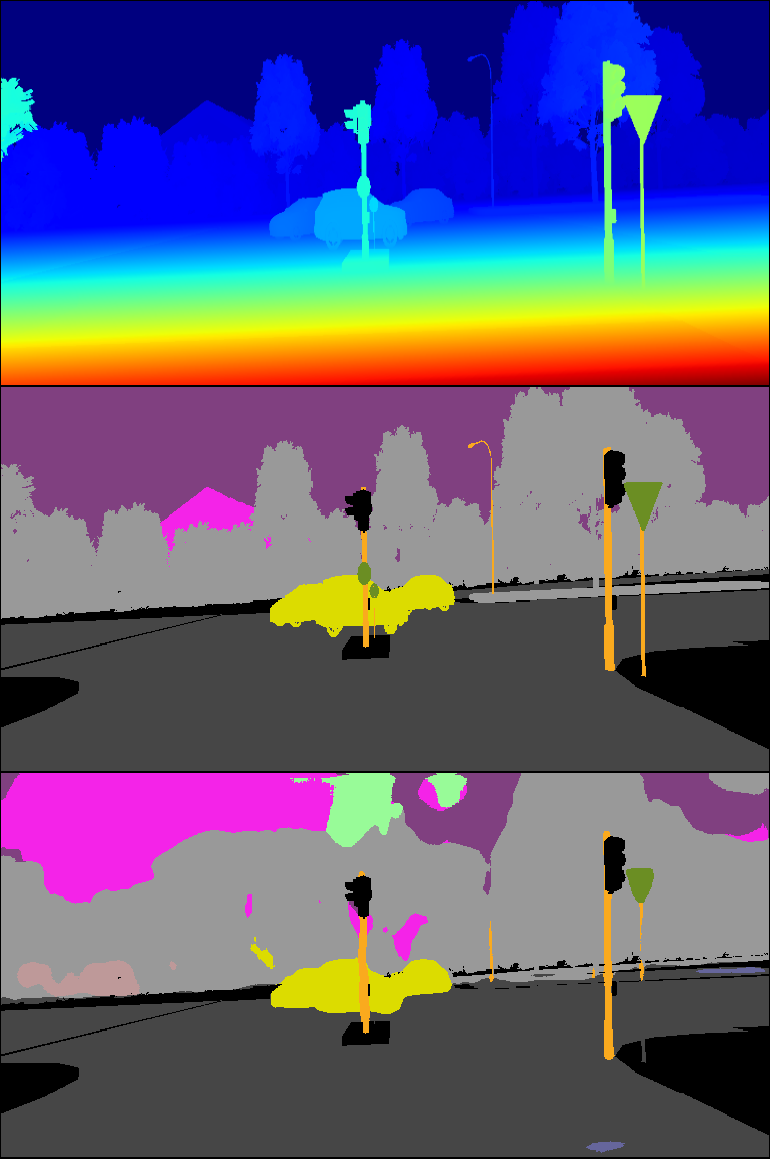}
\caption{Example RGB-depth pairs and segmentation results. Each image block shows (i) top: input frame (either RGB or depth); (ii) ground-truth; and (iii) predicted segmentation.}
\label{fig:ss_eg}
\end{figure}

Table~\ref{tab:ss_rgb} shows mAP of the Adapnet++ model trained on RGB images and tested on Virtual KITTI 2. One sees similar performance variations as for the earlier problems addressed. In particular, performance is relatively stable for small and moderate changes in the camera angle (15 and 30 degrees respectively) but degrades significantly in the presence of fog and rain.
Table~\ref{tab:ss_depth} shows results for Adapnet++ model trained on depth images. We first note that semantic segmentation performance is in general poorer when using depth images rather than RGB images, which is consistent with \cite{valada19ijcv}. Further, because Virtual KITTI 2's ground-truth depth is not affected by meteorological conditions or lighting, the depth images for the fog, morning, overcast, rain and sunset variations are identical, and therefore so is the performance of the model. As a point of comparison to real-world data, we evaluated the Adapnet++ RGB model on the 200 KITTI images with segmentation ground-truth. The mAP of 58.07 is in line with the average mAP across the 5 cloned scenes of 58.22. This indicates that Virtual KITTI 2 may be a good proxy dataset for evaluating segmentation algorithms on KITTI.
Figure~\ref{fig:ss_eg} shows examples of segmentation results for both RGB and depth frames. We can see that the RGB model performs well overall while the depth model often confuses the sky and building classes.

\begin{table}
\center
\caption{RGB semantic segmentation mAP results for different variations of Virtual KITTI 2.}
\begin{tabular}{llllll}
\rowcolor{gray!50}
condition & scene01 & scene02 & scene06 & scene18 & scene20 \\ 
\toprule
clone & 75.47 & 53.28 & 61.46 & 29.71 & 71.20 \\
fog & 44.89 & 34.75 & 43.39 & 46.80 & 48.11 \\
morning & 64.41 & 24.08 & 47.01 & 20.10 & 58.03 \\
overcast & 68.36 & 60.19 & 58.95 & 32.49 & 71.55 \\
rain & 48.77 & 44.74 & 45.15 & 39.56 & 52.47 \\
sunset & 59.59 & 47.90 & 55.22 & 30.95 & 51.37 \\
15-deg-left & 70.64 & 59.06 & 68.13 & 28.98 & 69.94 \\
15-deg-right & 74.05 & 46.73 & 60.37 & 31.61 & 71.63 \\
30-deg-left & 68.82 & 55.22 & 69.46 & 28.83 & 66.20 \\
30-deg-right & 70.49 & 44.63 & 57.11 & 30.72 & 71.03 \\
\bottomrule
\end{tabular}
\label{tab:ss_rgb}
\end{table}

\begin{table}
\center
\caption{Depth semantic segmentation mAP results for different variations of Virtual KITTI 2.}
\begin{tabular}{llllll}
\rowcolor{gray!50}
condition & scene01 & scene02 & scene06 & scene18 & scene20 \\ 
\toprule
clone & 44.57 & 34.75 & 40.84 & 37.56 & 34.24 \\
fog & 44.57 & 34.38 & 40.84 & 37.56 & 34.24 \\
morning & 44.57 & 34.75 & 40.84 & 37.56 & 34.24 \\
overcast & 44.57 & 34.38 & 40.84 & 37.56 & 34.24 \\
rain & 44.57 & 34.38 & 40.84 & 37.56 & 34.24 \\
sunset & 44.57 & 34.75 & 40.84 & 37.56 & 34.24 \\
15-deg-left & 42.86 & 32.99 & 43.40 & 40.40 & 30.82 \\
15-deg-right & 44.36 & 33.60 & 38.60 & 35.45 & 34.13 \\
30-deg-left & 37.82 & 30.97 & 40.07 & 39.32 & 26.18 \\
30-deg-right & 43.60 & 29.73 & 35.50 & 27.66 & 33.21 \\
\bottomrule
\end{tabular}
\label{tab:ss_depth}
\end{table}


\bibliographystyle{unsrt}  

\bibliography{references}
\end{document}